# Learning Continuous Time Bayesian Networks

**Uri Nodelman**
Stanford University
nodelman@cs.stanford.edu

**Christian R. Shelton**
Stanford University
cshelton@cs.stanford.edu

**Daphne Koller**
Stanford University
koller@cs.stanford.edu

## Abstract

Continuous time Bayesian networks (CTBN) describe structured stochastic processes with finitely many states that evolve over continuous time. A CTBN is a directed (possibly cyclic) dependency graph over a set of variables, each of which represents a finite state continuous time Markov process whose transition model is a function of its parents. We address the problem of learning parameters and structure of a CTBN from fully observed data. We define a conjugate prior for CTBNs and show how it can be used both for Bayesian parameter estimation and as the basis of a Bayesian score for structure learning. Because acyclicity is not a constraint in CTBNs, we can show that the structure learning problem is significantly easier, both in theory and in practice, than structure learning for dynamic Bayesian networks (DBNs). Furthermore, as CTBNs can tailor the parameters and dependency structure to the different time granularities of the evolution of different variables, they can provide a better fit to continuous-time processes than DBNs with a fixed time granularity.

## 1 Introduction

Learning about complex dynamic systems is an important task. From learning the timing and organization of metabolic pathways in cells to studying trends in demographic data to analyzing web server logs, there are many different processes that we would like to understand.

Dynamic Bayesian networks (DBNs) (Dean & Kanazawa, 1989) are a standard model used to learn and reason about dynamic systems. DBNs model a temporal process by discretizing time and providing a Bayesian network fragment that represents the probabilistic transition from the state at time $t$ to the state at time $t + \Delta t$. Using such a model for learning about the structure of a dynamic system has a significant limitation — namely, the structure we learn may be a function of the $\Delta t$ parameter we choose as much as it is a function of the underlying structure of the process.

The discretization of time imposes several problems. First, in standard DBN models, we must choose a uniform time granularity. However, in many real-world processes different variables can have very different time granularities, making any single choice of $\Delta t$ inappropriate. Second, when we discretize time, we aggregate into a variable's transition model all of the state changes that it takes over the entire course of the time slice. When the variable evolves at a finer time granularity than $\Delta t$, this approximation can be a very poor one.

Furthermore, the best possible time-sliced approximation to the variable's evolution model can be quite complex. First, it will often involve dependency edges within a time slice, which significantly complicates learning algorithms. Second, a DBN that best encodes the aggregated dependency will exhibit entanglement: As the discretization loses information about the values of a variable's parents, the values of the variable's ancestors might become relevant. Thus, a time-sliced DBN that represents the process dynamics will often be densely connected, which both obscures the true structure of the process, and makes it hard to learn from limited data.

In (Nodelman et al., 2002) we presented the alternative framework of *continuous time Bayesian networks (CTBNs)*. This framework allows for modelling stochastic processes over a structured state space evolving in continuous time. In this paper, we consider the problem of learning the structure of CTBNs from data. We use a Bayesian learning framework for CTBNs, using a Bayesian scoring function derived from an appropriate conjugate parameter prior. We provide an algorithm that searches over the space of possible CTBN structures for one that maximizes this Bayesian score; this algorithm is significantly simpler than DBN learning algorithms.

## 2 Continuous Time Bayesian Nets

In this section, we review the *continuous time Bayesian network (CTBN)* framework presented in (Nodelman et al., 2002). A CTBN represents a stochastic process over a structured state space, consisting of assignments to some set of local variables $\boldsymbol{X} = \{X_1, \ldots, X_n\}$, where each $X_i$ has a finite domain of values $Val(X_i)$.

### 2.1 Markov Processes

Let us first consider a Markov process over a single variable. A finite state, continuous time, homogeneous Markov process $X(t)$ with state space $Val(X) = \{x_1, \ldots, x_k\}$ is described by an initial distribution $P_X^0$ and an $n \times n$ matrix

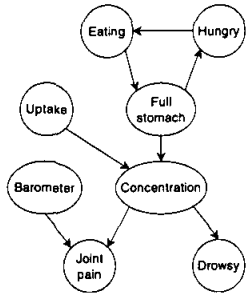

Figure 1: Drug effect network

of transition *intensities*:

$$\mathbf{Q}_X = \begin{bmatrix} -q_{x_1} & q_{x_1 x_2} & \cdots & q_{x_1 x_k} \\ q_{x_2 x_1} & -q_{x_2} & \cdots & q_{x_2 x_k} \\ \vdots & \vdots & \ddots & \vdots \\ q_{x_k x_1} & q_{x_k x_2} & \cdots & -q_{x_k} \end{bmatrix},$$

where $q_{x_i x_j}$ is the intensity of transitioning from state $x_i$ to state $x_j$ and $q_{x_i} = \sum_{j \neq i} q_{x_i x_j}$.

Given $\mathbf{Q}_X$ we can describe the transient behavior of $X(t)$ as follows. If $X(0) = x$ then it stays in state $x$ for an amount of time exponentially distributed with parameter $q_x$. Thus, the probability density function $f$ for $X(t)$ remaining at $x$ is $f(q_x, t) = q_x \exp(-q_x t)$ for $t \geq 0$, and the corresponding probability distribution function $F$ for $X(t)$ remaining at $x$ for an amount of time $\leq t$ is given by $F(q_x, t) = 1 - \exp(-q_x t)$. The expected time of transitioning is $1/q_x$. Upon transitioning, $X$ shifts to state $x'$ with probability $q_{xx'}/q_x$.

The distribution over transitions of $X$ factors into two pieces: an exponential distribution over *when* the next transition will occur and a multinomial distribution over *where* the state transitions — the next state of the system. The natural parameter for the exponential distribution is $q_x$ and the natural parameters for the multinomial distribution are $\theta_{xx'} = q_{xx'}/q_x, x' \neq x$. (As staying in the state $x$ is not a transition of $X$, there is no multinomial parameter $\theta_{xx}$.)

### 2.2 Continuous Time Bayesian Networks

We model the joint dynamics of several local variables by allowing the transition model of each local variable $X$ to be a Markov process whose parameterization depends on some subset of other variables $\mathbf{U}$. The key building block is a *conditional Markov process*:

**Definition 2.1** *A* conditional Markov process $X$ *is an inhomogeneous Markov process whose intensity matrix varies with time, but only as a function of the current values of a set of discrete conditioning variables* $\mathbf{U}$. *Its intensity matrix, called a* conditional intensity matrix *(CIM), is written* $\mathbf{Q}_{X|\mathbf{U}}$ *and can be viewed as a set of homogeneous intensity matrices* $\mathbf{Q}_{X|\mathbf{u}}$ *— one for each instantiation of values* $\mathbf{u}$ *to* $\mathbf{U}$. ∎

The parameters of $\mathbf{Q}_{X|\mathbf{U}}$ are $\boldsymbol{q}_{X|\mathbf{u}} = \{q_{x|\mathbf{u}} : x \in Val(X)\}$ and $\boldsymbol{\theta}_{X|\mathbf{u}} = \{\theta_{xx'|\mathbf{u}} : x, x' \in Val(X), x \neq x'\}$.

We can now combine a set of conditional Markov processes to form a CTBN:

**Definition 2.2** *A* continuous time Bayesian network $\mathcal{N}$ *over* $\mathbf{X}$ *consists of two components: an* initial distribution $P_{\mathbf{X}}^0$, *specified as a Bayesian network* $\mathcal{B}$ *over* $\mathbf{X}$, *and a* continuous transition model, *specified as*

- *A directed (possibly cyclic) graph* $\mathcal{G}$ *whose nodes are* $X_1, \ldots, X_n$; $Pa_{\mathcal{G}}(X_i)$, *often abbreviated* $\mathbf{U}_i$, *denotes the parents of* $X_i$ *in* $\mathcal{G}$.

- *A conditional intensity matrix,* $\mathbf{Q}_{X_i|\mathbf{U}_i}$, *for each variable* $X_i \in \mathbf{X}$. ∎

The learning problem for the initial distribution is a standard Bayesian network learning task, and we therefore ignore it for the remainder of this paper.

**Example 2.3** *Figure 1 shows the graph structure for a CTBN modelling the effect of a drug a person might take to alleviate pain in their joints. There are nodes for the uptake of the drug and for the resulting concentration of the drug in the bloodstream. The concentration is affected by how full the patient's stomach is. The pain may be aggravated by falling pressure. The drug may also cause drowsiness. The model contains a cycle, indicating that whether a person is hungry depends on how full their stomach is, which depends on whether or not they are eating, which in turn depends on whether they are hungry.*

The transitions of each local variable in a CTBN are controlled by the values of its parents. In the drug effect example above, when the concentration of the drug is *low* and barometric pressure is *falling*, the transition model of the variable *JointPain* is a Markov process parameterized by $\mathbf{Q}_{JointPain|low,falling}$; in this model a transition to the value *JointPain=high* is likely. When the concentration of the drug rises to *high* (due to drug uptake), the parameterization of the transition model of *JointPain* changes to $\mathbf{Q}_{JointPain|high,falling}$, in which a transition to the value *JointPain=high* is much less likely.

## 3 CTBN Parameter Estimation

We first consider the problem of estimating the parameters of a CTBN with a fixed structure $\mathcal{G}$. As usual, this problem is not only useful on its own, but also as a key component in the structure learning task.

Our data are a set of trajectories $D = \{\sigma_1, \ldots, \sigma_h\}$ where each $\sigma_i$ is a complete set of state transitions and the times at which they occurred. So, for each point in time, we know the full instantiation to all variables.

### 3.1 The Likelihood Function

As in any density estimation task, a key element is the likelihood function.

**Single Markov Process.** Consider a homogeneous Markov process $X(t)$. As all the transitions are observed, the likelihood of $D$ can be decomposed as a product of the likelihoods for individual transitions $d$. Let $d = \langle x_d, t_d, x'_d \rangle \in D$ be the transition where $X$ transitions to state $x'_d$ after spending the amount of time $t_d$ in state $x_d$. We can write the likelihood for the single transition $d$ as

$$L_X(q, \theta : d) = \underbrace{(q_{x_d} \exp(-q_{x_d} t_d))}_{L_X(q:d)} \underbrace{\left(\theta_{x_d x'_d}\right)}_{L_X(\theta:d)}.$$

By multiplying the likelihoods for each transition $d$, we see that we can summarize our data $D$ in terms of the sufficient statistics $T[x]$, the amount of time $X$ spends in state $x$, and $M[x, x']$, the number of times $X$ transitions from $x$ to $x'$ for $x \neq x'$. If we write $M[x] = \sum_{x'} M[x, x']$, the total number of transitions leaving the state $X = x$ then we have

$$L_X(q, \theta : D) = \left(\prod_{d \in D} L(q : d)\right) \left(\prod_{d \in D} L_X(\theta : d)\right)$$

$$= \left(\prod_x q_x^{M[x]} \exp(-q_x T[x])\right) \left(\prod_x \prod_{x' \neq x} \theta_{xx'}^{M[x,x']}\right). \quad (1)$$

**CTBNs.** In a CTBN $\mathcal{N}$, each variable $X \in \mathbf{X}$ is conditioned on its parent set $\mathbf{U}$, and each transition of $X$ is considered in the context of the instantiation to $\mathbf{U}$. With complete data, we know the value of $\mathbf{U}$ during the entire trajectory, so we know at each point in time precisely which homogeneous intensity matrix $\mathbf{Q}_{X|\mathbf{u}}$ governed the dynamics of $X$. Thus, the likelihood decomposes by variable as

$$L_{\mathcal{N}}(q, \theta : D) = \prod_{X_i \in \mathbf{X}} L_{X_i}(q_{X_i|\mathbf{U}_i}, \theta_{X_i|\mathbf{U}_i} : D)$$

$$= \prod_{X_i \in \mathbf{X}} L_{X_i}(q_{X_i|\mathbf{U}_i} : D) L_{X_i}(\theta_{X_i|\mathbf{U}_i} : D).$$

The term $L_X(\theta_{X|\mathbf{U}} : D)$ is the probability of the sequence of state transitions, disregarding the times between transitions. These state changes depend only on the value of the parents at the instant of the transition. Therefore, with the sufficient statistic $M[x, x'|\mathbf{u}]$, we have

$$L_X(\theta : D) = \prod_{\mathbf{u}} \prod_x \prod_{x' \neq x} \theta_{xx'|\mathbf{u}}^{M[x,x'|\mathbf{u}]}.$$

The computation of $L_X(q_{X|\mathbf{U}} : D)$ is more subtle. Consider a particular transition $d$ where a state in which $X = x, \mathbf{U} = \mathbf{u}$ transitioned to another state $X = x, \mathbf{U} = \mathbf{u}'$ after time $t$. In other words, the duration in the state was terminated not due to a transition of $X$, but due to a transition of one of its parents. Intuitively, these transitions still depend on $X$'s dynamics, as they can only occur if $X$ stayed at the value $x$ for at least a duration of $t$. The probability that $X$ stayed at $x$ for this duration is $1 - F(q_{x|\mathbf{u}}, t) = \exp(-q_{x|\mathbf{u}} t)$.

More formally, the sufficient statistic $T[x|\mathbf{u}]$, the total amount of time where $X = x$ and $\mathbf{U} = \mathbf{u}$, can be decomposed into two different kinds of durations: $T[x|\mathbf{u}] = T_r[x|\mathbf{u}] + T_c[x|\mathbf{u}]$, where $T_r[x|\mathbf{u}]$ is the total over durations $t_d$ that terminate with $X$ remaining equal to $x$ (these include transitions where $\mathbf{U}$ changed value, as well as the end of a trajectory), and $T_c[x|\mathbf{u}]$ is the total over durations $t_d$ that terminate with a change in the value of $X$. It is easy to see that the terms for the different transitions that comprise $T_r[x|\mathbf{u}]$ combine, so that we have

$$L_X(q_{X|\mathbf{U}} : D) = \left(\prod_{\mathbf{u}} \prod_x q_{x|\mathbf{u}}^{M[x|\mathbf{u}]} \exp(-q_{x|\mathbf{u}} T_c[x|\mathbf{u}])\right)$$

$$\times \left(\prod_{\mathbf{u}} \prod_x \exp(-q_{x|\mathbf{u}} T_r[x|\mathbf{u}])\right)$$

$$= \prod_{\mathbf{u}} \prod_x q_{x|\mathbf{u}}^{M[x|\mathbf{u}]} \exp(-q_{x|\mathbf{u}} T[x|\mathbf{u}]).$$

Thus, we do not need to maintain the distinction between $T_c[x|\mathbf{u}]$ and $T_r[x|\mathbf{u}]$. Instead, we can simply use $T[x|\mathbf{u}]$ as the sufficient statistic.

We can now write the log likelihood as a sum of local variable likelihoods of the form

$$\ell_X(q, \theta : D) = \ell_X(q : D) + \ell_X(\theta : D)$$

$$= \left[\sum_{\mathbf{u}} \sum_x M[x|\mathbf{u}] \ln(q_{x|\mathbf{u}}) - q_{x|\mathbf{u}} \cdot T[x|\mathbf{u}]\right]$$

$$+ \left[\sum_{\mathbf{u}} \sum_x \sum_{x' \neq x} M[x, x'|\mathbf{u}] \ln(\theta_{xx'|\mathbf{u}})\right].$$

From this formula, we can derive the MLE parameters:

$$\hat{q}_{x|\mathbf{u}} = \frac{M[x|\mathbf{u}]}{T[x|\mathbf{u}]}; \quad \hat{\theta}_{xx'|\mathbf{u}} = \frac{M[x, x'|\mathbf{u}]}{M[x|\mathbf{u}]}. \quad (2)$$

### 3.2 The Bayesian Approach

To perform Bayesian parameter estimation, and to define a Bayesian score for our structure search, we need to define a prior distribution over the parameters of our CTBN. As usual, for computational efficiency, we want to use a *conjugate prior* — one where the posterior (after conditioning on the data) is in the same parametric family as the prior.

Let us begin with constructing an appropriate prior for a single Markov process. Recall that a Markov process has two sets of parameters: a multinomial distribution parameterized by $\theta$, and an exponential distribution parameterized by $q$. The multinomial distribution is familiar from traditional Bayesian networks where the standard conjugate prior is a Dirichlet distribution (Heckerman et al., 1995; Geiger & Heckerman, 1995). An appropriate conjugate prior for the exponential parameter $q$ is the *Gamma distribution* $P(q) = Gamma(\alpha, \tau)$, where

$$P(q) = \frac{(\tau)^{\alpha+1}}{\Gamma(\alpha + 1)} q^\alpha \exp(-q\tau).$$

If we assume that

$$P(\boldsymbol{\theta}) = Dir(\alpha_{xx_1}, \ldots, \alpha_{xx_k})$$
$$P(q) = Gamma(\alpha_x, \tau_x)$$
$$P(\boldsymbol{\theta}, q) = P(\boldsymbol{\theta})P(q),$$

then, after conditioning on the data, we have

$$P(\boldsymbol{\theta} \mid D) = Dir(\alpha_{xx_1} + M[x, x_1], \ldots, \alpha_{xx_k} + M[x, x_k])$$
$$P(q \mid D) = Gamma(\alpha_x + M[x], \tau_x + T[x]) \ .$$

We generalize this idea to a parameter prior for an entire CTBN by making two standard assumptions for parameter priors in Bayesian networks (Heckerman et al., 1995). *global parameter independence*:

$$P(\boldsymbol{q}, \boldsymbol{\theta}) = \prod_{X \in \mathbf{X}} P(q_{X|\mathbf{Pa}(X)}, \boldsymbol{\theta}_{X|\mathbf{Pa}(X)})$$

and *local parameter independence*:

$$P(\boldsymbol{q}_{X|\mathbf{U}}, \boldsymbol{\theta}_{X|\mathbf{U}}) = \left(\prod_{\mathbf{u}} P(q_{x|\mathbf{u}})\right) \left(\prod_{x} \prod_{\mathbf{u}} P(\boldsymbol{\theta}_{x|\mathbf{u}})\right) \ .$$

If our parameter prior satisfies these assumptions, so does our posterior, as it belongs to the same parametric family. Thus, we can maintain our parameter distribution in closed form, and update it using the obvious sufficient statistics: $M[x, x'|\mathbf{u}]$ for $\boldsymbol{\theta}_{x|\mathbf{u}}$, and $M[x|\mathbf{u}], T[x|\mathbf{u}]$ for $q_{x|\mathbf{u}}$.

Given a parameter distribution, we can use it to predict the next event, averaging out the event probability over the possible values of the parameters. As usual, this prediction is equivalent to using "expected" parameter values, which have the same form as the MLE parameters, but account for the "imaginary counts" of the hyperparameters:

$$\hat{q}_{x|\mathbf{u}} = \frac{\alpha_{x|\mathbf{u}} + M[x|\mathbf{u}]}{\tau_{x|\mathbf{u}} + T[x|\mathbf{u}]}; \quad \hat{\theta}_{xx'|\mathbf{u}} = \frac{\alpha_{xx'|\mathbf{u}} + M[x, x'|\mathbf{u}]}{\alpha_{x|\mathbf{u}} + M[x|\mathbf{u}]} \ . \quad (3)$$

Note that, in principle, this choice of parameters is only valid for predicting a single transition, after which we should update our parameter distribution accordingly. However, as is often done in other settings, we can "freeze" the parameters to these expected values, and use them for predicting an entire trajectory.

## 4 Learning CTBN Structure

We now turn to the problem of learning the structure of a CTBN. We take a score-based approach to this task, defining a Bayesian score for evaluating different candidate structures, and then using a search algorithm to find a structure that has high score.

### 4.1 Scoring CTBNs

The *Bayesian score* over structures $\mathcal{G}$ is defined as

$$\text{score}_B(\mathcal{G} : D) = \ln P(D \mid \mathcal{G}) + \ln P(\mathcal{G}) \ . \quad (4)$$

We can significantly increase the efficiency of our search algorithm if we assume that our prior satisfies certain standard assumptions. We assume that our structure prior $P(\mathcal{G})$ satisfies *structure modularity*, so that $P(\mathcal{G}) = \prod_i P(\mathbf{Pa}(X_i) = \mathbf{Pa}_{\mathcal{G}}(X_i))$. We also assume that our parameter prior satisfies *parameter modularity*: For any two structures $\mathcal{G}$ and $\mathcal{G}'$ such that $\mathbf{Pa}_{\mathcal{G}}(X) = \mathbf{Pa}_{\mathcal{G}'}(X)$, we have that $P(q_X, \boldsymbol{\theta}_X \mid \mathcal{G}) = P(q_X, \boldsymbol{\theta}_X \mid \mathcal{G}')$. Combining parameter modularity and parameter independence, we have

$$P(\boldsymbol{q}_{\mathcal{G}}, \boldsymbol{\theta}_{\mathcal{G}} \mid \mathcal{G}) = \prod_{X_i} P(q_{X_i|\mathbf{U}_i} \mid \mathbf{Pa}(X_i) = \mathbf{Pa}_{\mathcal{G}}(X_i))$$
$$P(\boldsymbol{\theta}_{X_i|\mathbf{U}_i} \mid \mathbf{Pa}(X_i) = \mathbf{Pa}_{\mathcal{G}}(X_i)).$$

As $P(\mathcal{G})$ does not grow with the amount of data, the significant term in Eq. (4) is the *marginal likelihood* $P(D \mid \mathcal{G})$. This term incorporates our uncertainty over the parameters by integrating over all of their possible values:

$$P(D \mid \mathcal{G}) = \int_{\boldsymbol{q}_{\mathcal{G}}, \boldsymbol{\theta}_{\mathcal{G}}} P(D \mid \boldsymbol{q}_{\mathcal{G}}, \boldsymbol{\theta}_{\mathcal{G}}) P(\boldsymbol{q}_{\mathcal{G}}, \boldsymbol{\theta}_{\mathcal{G}} \mid \mathcal{G}) d\boldsymbol{q}_{\mathcal{G}} d\boldsymbol{\theta}_{\mathcal{G}}.$$

As in Eq. (1), the likelihood decomposes as a product:

$$P(D \mid \boldsymbol{q}_{\mathcal{G}}, \boldsymbol{\theta}_{\mathcal{G}}) = \prod_{X_i} L_{X_i}(q_{X_i|\mathbf{U}_i} : D) L_{X_i}(\boldsymbol{\theta}_{X_i|\mathbf{U}_i} : D)$$
$$= \underbrace{\left(\prod_{X_i} L_{X_i}(q_{X_i|\mathbf{U}_i} : D)\right)}_{L(q:D)} \underbrace{\left(\prod_{X_i} L_{X_i}(\boldsymbol{\theta}_{X_i|\mathbf{U}_i} : D)\right)}_{L(\boldsymbol{\theta}:D)} \ .$$

Using this decomposition, and global parameter independence, we now have

$$P(D \mid \mathcal{G}) = \int_{\boldsymbol{q}_{\mathcal{G}}, \boldsymbol{\theta}_{\mathcal{G}}} L(\boldsymbol{q}_{\mathcal{G}} : D) L(\boldsymbol{\theta} : D) P(\boldsymbol{\theta}_{\mathcal{G}}) P(\boldsymbol{q}_{\mathcal{G}}) d\boldsymbol{q}_{\mathcal{G}} d\boldsymbol{\theta}_{\mathcal{G}}$$
$$= \left(\int_{\boldsymbol{q}_{\mathcal{G}}} L(\boldsymbol{q}_{\mathcal{G}} : D) P(\boldsymbol{q}_{\mathcal{G}}) d\boldsymbol{q}_{\mathcal{G}}\right) \quad (5)$$
$$\times \left(\int_{\boldsymbol{\theta}_{\mathcal{G}}} L(\boldsymbol{\theta}_{\mathcal{G}} : D) P(\boldsymbol{\theta}_{\mathcal{G}}) d\boldsymbol{\theta}_{\mathcal{G}}\right) \ . \quad (6)$$

Using local parameter independence, the term (5) can be decomposed as

$$\prod_{X \in \mathbf{X}} \prod_{\mathbf{u}} \prod_{x} \int_0^{\infty} P(q_{x|\mathbf{u}}) L_X(q_{x|\mathbf{u}} : D) dq_{x|\mathbf{u}}$$
$$= \prod_{X \in \mathbf{X}} \prod_{\mathbf{u}} \prod_{x} \int_0^{\infty} \frac{(\tau_{x|\mathbf{u}})^{\alpha_{x|\mathbf{u}}+1}}{\Gamma(\alpha_{x|\mathbf{u}}+1)} (q_{x|\mathbf{u}})^{\alpha_{x|\mathbf{u}}} \exp(-q_{x|\mathbf{u}}\tau_{x|\mathbf{u}})$$
$$\qquad \times (q_{x|\mathbf{u}})^{M[x|\mathbf{u}]} \exp(-q_{x|\mathbf{u}} \cdot T[x|\mathbf{u}]) dq_{x|\mathbf{u}}$$
$$= \prod_{X \in \mathbf{X}} \prod_{\mathbf{u}} \prod_{x} \int_0^{\infty} \left[\frac{(\tau_{x|\mathbf{u}})^{\alpha_{x|\mathbf{u}}+1}}{\Gamma(\alpha_{x|\mathbf{u}}+1)} (q_{x|\mathbf{u}})^{\alpha_{x|\mathbf{u}}+M[x|\mathbf{u}]}\right.$$
$$\qquad \left. \times \exp(-q_{x|\mathbf{u}}(\tau_{x|\mathbf{u}} + T[x|\mathbf{u}]))\right] dq_{x|\mathbf{u}}$$
$$= \prod_{X \in \mathbf{X}} \prod_{\mathbf{u}} \prod_{x} \frac{\Gamma(\alpha_{x|\mathbf{u}} + M[x|\mathbf{u}] + 1)(\tau_{x|\mathbf{u}})^{\alpha_{x|\mathbf{u}}+1}}{\Gamma(\alpha_{x|\mathbf{u}}+1)(\tau_{x|\mathbf{u}} + T[x|\mathbf{u}])^{\alpha_{x|\mathbf{u}}+M[x|\mathbf{u}]+1}}$$
$$= \prod_{X \in \mathbf{X}} MargL^q(X, \mathbf{Pa}_{\mathcal{G}}(X) : D) \ .$$

As the distributions over the parameters $\theta$ are Dirichlet, the analysis of the term Eq. (6) is analogous to traditional Bayesian networks, simplifying to

$$\prod_{X \in X} \prod_{\mathbf{u}} \prod_{x} \frac{\Gamma(\alpha_{x|\mathbf{u}})}{\Gamma(\alpha_{x|\mathbf{u}} + M[x|\mathbf{u}])}$$
$$\times \prod_{x' \neq x} \frac{\Gamma(\alpha_{xx'|\mathbf{u}})}{\Gamma(\alpha_{xx'|\mathbf{u}} + M[x,x'|\mathbf{u}])}$$
$$= \prod_{X \in X} MargL^\theta(X, \mathbf{Pa}_{\mathcal{G}}(X) : D) .$$

Using this decomposition, and the assumption of structure modularity, the Bayesian score in Eq. (4) can now be decomposed as a sum of family scores **FamScore**$(X, \mathbf{Pa}_{\mathcal{G}}(X) : D)$ that measures the quality of $\mathbf{Pa}_{\mathcal{G}}(X)$ as a parent set for $X$ given data $D$:

$$\text{score}_B(\mathcal{G} : D) = \sum_{X_i \in X} \textbf{FamScore}(X_i, \mathbf{Pa}_{\mathcal{G}}(X_i) : D)$$
$$= \sum_{X_i \in X} \ln P(\mathbf{Pa}(X) = \mathbf{Pa}_{\mathcal{G}}(X_i)) +$$
$$\ln MargL^q(X_i, \mathbf{U}_i : D) + \ln MargL^\theta(X_i, \mathbf{U}_i : D) .$$

### 4.2 Model Search

Given the score function, it remains to find a structure $\mathcal{G}$ that maximizes the score. This task is an optimization problem over possible CTBN network structures. Interestingly, the search space over CTBN structures is significantly simpler than that of BNs or DBNs.

Chickering et al. (1994) show that the problem of learning a optimal Bayesian network structure is NP-hard. Specifically, they define the problem **k-Learn**: Finding the highest scoring Bayesian network structure, when each variable is restricted to have at most $k$ parents. The problem **k-Learn** is NP-hard even for $k = 2$. Intuitively, the reason is that we cannot determine the optimal parent set for each node individually; due to the acyclicity constraint, the choice of parent set for one node restricts our choices for other nodes. The same NP-hardness result clearly carries over to DBNs, if we allow edges within a time slice.

However, this problem does not arise in the context of CTBN learning. Here, all edges are across time — representing the effect of the current value of one variable on the next value of the other. Thus, we have no acyclicity constraints, and we can optimize the parent set for each variable independently. Specifically, if we restrict the maximum number of parents to $k$, we can simply exhaustively enumerate each of the possible parent sets $\mathbf{U}$ for $|\mathbf{U}| \leq k$ and compute **FamScore**$(X \mid \mathbf{U} : D)$. We then choose as $Pa(X)$ the set $\mathbf{U}$ which maximizes the family score. For fixed $k$, this algorithm is polynomial in $n$. Therefore,

**Theorem 4.1** *The problem **k-Learn** for CTBNs, for fixed $k$, can be solved in polynomial time in the number of variables $n$ and the size of the dataset $D$.*

In practice, we do not wish to exhaustively enumerate the possible parent sets for each variable $X$. We can therefore use a greedy hill-climbing search with operators that add and delete edges in the CTBN graph. However, due to the lack of interactions between the families of different variables, we can perform this greedy search separately for each variable $X$, selecting a locally optimal family for it. Thus, this heuristic search can be performed much more efficiently than for BNs or DBNs.

## 5 Structure Identifiability

So far, we have focused on the problem of learning a CTBN that provides a good fit to some training data $D$. However, we have not addressed the fundamental question of the scope of this learning procedure: Which stochastic processes can we represent using a CTBN, and can we reliably identify them from training data?

### 5.1 Representational Ability

We begin by considering the scope of the CTBN representation: Which underlying distributions can we represent using a CTBN? More formally, we say that two Markov processes are *stochastically equivalent* if they have the same state space and transition probabilities (Gihman & Skorohod, 1973). Now, consider a homogeneous stochastic process over *Val*($X$), defined as an intensity matrix $\mathbf{Q}_X$. We would like to determine when there is a CTBN which is stochastically equivalent to $\mathbf{Q}_X$.

In Nodelman et al. (2002), we provided a semantics for a CTBN in terms of an *amalgamation* operation, which takes a CTBN and converts it into a single intensity matrix that specifies a homogeneous stochastic process. For a CTBN $\mathcal{N}$, let $\mathbf{Q}^\mathcal{N}$ be the induced joint intensity matrix. We can now define

**Definition 5.1** *A CTBN structure $\mathcal{G}$ is an S-map for a homogeneous stochastic process $\mathbf{Q}_X$ if there exists a CTBN $\mathcal{N}$ over the graph $\mathcal{G}$ such that $\mathbf{Q}^\mathcal{N}$ is stochastically equivalent to $\mathbf{Q}_X$.*

As discussed in Nodelman et al. (2002), a basic assumption in the semantics of CTBNs is that, as time is continuous, variables cannot transition at the same instant. Thus, in the joint intensity matrix, all intensities that correspond to two simultaneous changes are zero. More precisely:

**Definition 5.2** *A homogeneous stochastic process $\mathbf{Q}_X$ with entries $q_{xx'}$ is said to be variable-based if, for any two assignments $x$ and $x'$ to $X$ that differ on more than one variable, $q_{xx'} = 0$.*

It turns out that this condition is the only restriction on the CTBN expressive power. Let $\mathcal{G}^\top$ be the fully connected directed graph. Then we can show that

**Theorem 5.3** *The graph $\mathcal{G}^\top$ is a S-map for any variable-based homogeneous stochastic process $\mathbf{Q}_X$.*

Thus, we can represent every variable-based homogeneous process as some parameterization over the graph $\mathcal{G}^\top$. In fact, this parameterization is unique:

**Theorem 5.4** *Let $\mathcal{N}$ and $\mathcal{N}'$ be two CTBNs over $\mathcal{G}^\top$. Then $\mathbf{Q}^\mathcal{N}$ and $\mathbf{Q}^{\mathcal{N}'}$ are stochastically equivalent if and only if their conditional intensity matrices are identical.*

Let $\mathcal{N}_{\mathbf{Q}_X}$ represent the unique CTBN over $\mathcal{G}^\top$ which is stochastically equivalent to $\mathbf{Q}_X$.

Although capturing a stochastic process using a fully-connected CTBN is not very interesting, it provides us with the tools for proving our main result.

**Theorem 5.5** *A CTBN structure $\mathcal{G}$ is an S-map for a variable-based process $\mathbf{Q}_X$ if and only if $\mathcal{N}_{\mathbf{Q}_X}$ satisfies the following condition: For every variable $X$, and any two assignments $x, x'$ to $\mathrm{Val}(X)$ such that $x$ and $x'$ agree on the value of $X$ and $\mathbf{Pa}_\mathcal{G}(X)$, we have that $q_x = q_{x'}$.*

Thus, we cannot represent the same process using two fundamentally different CTBN structures. We can only add spurious edges, corresponding to vacuous dependencies.

**Theorem 5.6** *For any variable-based process $\mathbf{Q}_X$, there exists a structure $\mathcal{G}^*$ such that, for any S-map $\mathcal{G}$ for $\mathbf{Q}_X$, $\mathcal{G}^* \subseteq \mathcal{G}$.*

Let us compare this result to the case of Bayesian networks. There, any distribution has many minimal I-maps; indeed, many distributions even have several perfect maps, each of which captures the structure of the distribution perfectly. In the case of CTBNs, we have a unique minimal S-map. To obtain some intuition for this difference, consider the simple example of a two-variable CTBN $\mathcal{N}$ with the graph $X \to Y$. Unless the edge between $X$ and $Y$ is vacuous, this graph cannot give rise to the same transition probabilities as any CTBN $\mathcal{N}'$ with the graph $X \leftarrow Y$. To see that, recall that in $\mathcal{N}$, the variable $Y$ is an inhomogeneous Markov process whose transition probabilities vary over time as a function of the changing value of $X$. But, in $\mathcal{N}'$, the variable $Y$ is a homogeneous Markov process whose transition probabilities never change.

### 5.2 Identifiability

Now that we have determined that any variable-based stochastic process has a unique minimal CTBN representation, the main question is whether we can identify this CTBN from data. More precisely, assume that our data $D$ is generated from some process $\mathbf{Q}_X$, and let $\mathcal{G}^*$ be the minimal S-map for $\mathbf{Q}_X$. We would like our learning algorithm to return a network whose structure is $\mathcal{G}^*$. Our learning algorithm searches for the network structure that maximizes the Bayesian score. Thus, the key property (ignoring possible limitations of our search procedure) is the following.

**Definition 5.7** *A scoring function is said to be consistent if, as the amount of data $|D| \to \infty$, the following holds with probability that approaches 1: The structure $\mathcal{G}^*$ will maximize the score, and the score of all structures $\mathcal{G} \neq \mathcal{G}^*$ will have a strictly lower score.*

Once again, compare this situation to that of Bayesian networks. There, the best we can hope for is that all and only structures that are I-equivalent to the "true" network will maximize the score.

To prove that our score is consistent, it helps to consider its behavior as the amount of data increases.

**Theorem 5.8** *As the amount of data $|D| \to \infty$,*

$$score_B(\mathcal{G} : D) = \ell(\hat{q}_\mathcal{G}, \hat{\theta}_\mathcal{G} : D) - \frac{\ln|D|}{2}Dim[\mathcal{G}] + O(1) \quad (7)$$

*where $Dim[\mathcal{G}]$ is the number of independent parameters in $\mathcal{G}$, and $\hat{q}_\mathcal{G}$ and $\hat{\theta}_\mathcal{G}$ are the MLE parameters of Eq. (2).*

Eq. (7) is simply the standard BIC approximation to the Bayesian score (Lam & Bacchus, 1994), which carries over to CTBNs. It shows that, asymptotically, the CTBN Bayesian score trades off fit to data and model complexity. We are more likely to add an arc if it represents a strong connection between the variables. Moreover, as the amount of data grows, we obtain more support for weak connections, and are more likely to introduce additional arcs.

**Theorem 5.9** $score_B(\mathcal{G} : D)$ *is consistent.*

The proof shows that the BIC score is consistent; as consistency is an asymptotic property, it suffices to show the consistency of the Bayesian score. The argument for the consistency of the BIC score is a standard one: If $\mathcal{G}$ is a superset of $\mathcal{G}^*$, it can represent $\mathbf{Q}_X$ exactly; thus, with enough data, the difference between the log-likelihood components of the score of $\mathcal{G}$ and $\mathcal{G}^*$ will go to zero. But, $\mathcal{G}$ has more parameters, leading to a higher penalty and thus a lower score. If $\mathcal{G}$ is not a superset of $\mathcal{G}^*$, it follows from Theorem 5.6 that it is not capable of representing $\mathbf{Q}_X$. In this case, as the amount of data grows, the likelihood portion of the score will dominate and $\mathcal{G}^*$ will have the higher score.

## 6 Experimental Results

We tested our CTBN learning framework on various synthetic data sets, generated from CTBNs. We used a simple greedy hill-climbing algorithm over the space of structures, optimizing the family for each variable separately. For comparison, we also learned DBNs using different time granularities. To allow a fair comparison, we used the same greedy hill-climbing algorithm there.

We first tested ability of CTBNs and DBNs to capture very simple dependencies. We constructed a CTBN model with four binary variables arranged in a chain. The first variable randomly switches between its states on a time scale of 1 time unit and each of the other variables follows its predecessor on the same time scale. In total, there are 14 parameters in this network. We learned a CTBN structure with increasing amounts of data, and DBNs with with varying time granularities. The number of parameters learned can be seen in Figure 2(a). The CTBN learning converges very quickly to the correct number of parameters, and, indeed, to the correct structure. Moreover, as we can see from the error bars, there is very low variance in the structures produced. By contrast, the DBN learning algorithm fluctuates significantly, and does not converge to the right number of parameters even with a large amount of data.

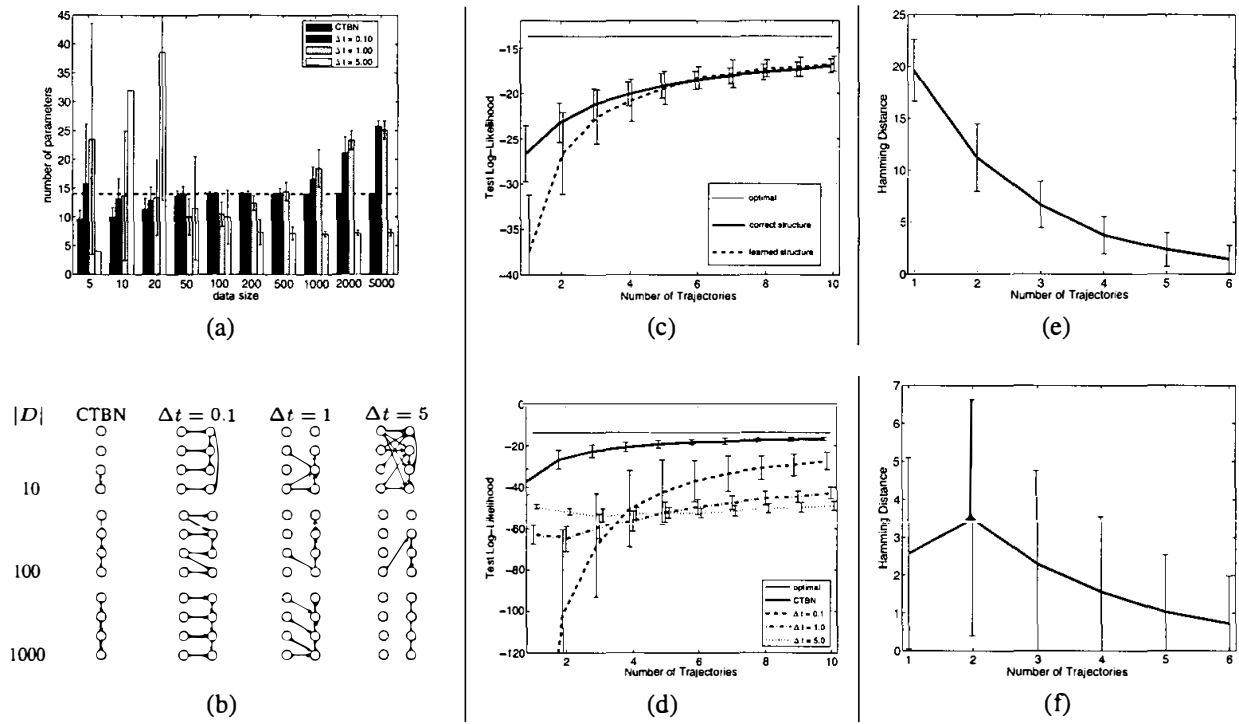

Figure 2: (a) For a 4-node chain network, the number of parameters of the learned structures as a function of the amount of time the data was collected, for CTBNs and DBNs with varying time granularity. (b) Example learned structures for the 4-node chain network. (c) & (d) Log-likelihoods of test data for networks learned from varying amounts of data generated from the drug effect network. Each trajectory corresponds to 6 units of time, and about 18 transitions. The thin line shows the likelihood for the true network. (c) CTBN with learned parameters and structure and CTBN with learned parameters only. (d) Learned CTBN model and DBN models with differing time slice durations. (e) & (f) Hamming distances for randomly generated 10-node CTBNs, for varying amounts of data. Each trajectory corresponds to 150 units of time, and about 1000 transitions. (e) Distance between the true structure and the highest scoring structure. (f) Distance between the highest scoring structure and the structure learned by greedy search for random CTBNs.

Typical structures are shown in Figure 2(b). As we can see, a DBN with time slices much shorter than the average rate of change of the system does converge to a reasonable structure; however, for large amounts of data, the structure still becomes more complex than the corresponding CTBN. For a time granularity on the same order as the time scale of the system, things become more difficult for the DBN, as it must model multiple transitions in a single time step, leading to entanglement which increases with the amount of available data. Finally, if the time-slicing is too coarse, the DBN learns a model of the steady-state distribution without any model of the transition probabilities.

We then tested our ability to recover more complex structures. We generated different amounts of data from the drug effect network of Figure 1, and used it to learn two models: one where we learned both the CTBN structure and the parameters, and the other where we simply estimated parameters for the correct network structure. We then computed the log-likelihood of test data for all networks, including the generating network. In all cases where we used a learned network, we used the expected parameters of Eq. (3) throughout the entire test trajectory. The results are shown in Figure 2(c). Even for fairly small amounts of data, our results with unknown structure are essentially identical to those with the correct structure.

To further test the ability of our algorithm to recover structure, we generated 100 random networks of 10 binary processes. We fixed a maximum parent set size of 4 and generated a random graph structure obeying this constraint. We then drew the multinomial parameters of the network from Dirichlet distributions (with parameters all 1) and the exponential parameters from a Gamma distribution (with both parameters equal to 1). In figure 2(e) we compared the maximum-score structure (with the same constraint on parent sets) to the true structure. The Hamming distance measured is the number of arcs present in only one of the graphs. As predicted by Theorem 5.9, as the amount of data grows, the correct structure has the highest score. Indeed, this happens even for very reasonable amounts of data. More interestingly, in a very large fraction of the cases, the simple greedy search algorithm recovers the highest scoring network very reliably. Figure 2(f) shows the difference between the maximum-score structure and the one found by greedy search. As we can see, the local minima in the search space are less frequent as the amount of data grows, and in general the difference between the exhaustive and greedy search techniques is small (roughly one edge difference for reasonable amounts of data).

Finally, we wanted to compare the generalization performance of learned CTBNs with those of learned DBNs. To do so, we had to extend the DBN model to include distributions over when, within a time slice, a given transition occurred. We assumed a uniform distribution within the time slice, augmenting it with a parameter for each variable that determines the probability that the value of the variable transitions more than once within a time slice. The value of this parameter was also learned from data. Figure 2(d) compares the generalization ability of learned CTBNs and learned DBNs with varying time granularity. As expected, the correct DBN structure exhibits entanglement due to the temporal discretization, and therefore requires more edges to approximate the distribution well. Even for small $\Delta t$, the amount of data required to estimate the much larger number of parameters is significantly greater. As $\Delta t$ grows large, the performance of the DBN decreases rapidly. Interestingly, as in the chain network, for large values of $\Delta t$, the DBNs simply cannot capture the transition dynamics accurately enough to converge to competitive performance.

## 7 Discussion and Conclusions

We have presented a Bayesian structure learning algorithm for continuous time Bayesian networks. As we showed, learning temporal processes as a CTBN has several important advantages. As we are not discretizing time, we do not need to choose some single time granularity in which to model the process. The model for each variable can reflect its own time granularity, better representing its evolution. Furthermore, as CTBNs do not aggregate multiple transitions over the course of a time slice, they avoid entanglement due to aggregation. Thus, they allow us to learn a model that more directly reflects the dependencies in the process. Finally, with no intra-time-slice edges, acyclicity is not a concern, so that the task of searching for a high-scoring network is computationally significantly simpler, both in theory and in practice, than in the case of DBNs.

It is useful to compare the ability of DBNs and CTBNs to represent different temporal processes. CTBNs are designed to represent purely Markovian processes — those where the instantaneous transition model depends only on the current state. Such processes can be represented very compactly as a CTBN, taking full advantage of any structure. By contrast, to represent these dynamics correctly as a DBN, we would need to aggregate the influence of one variable on another over the entire time slice, leading to entanglement of the influences and thereby to a more complicated network structure, with more parameters, that obscures the independencies in the underlying process. But, DBNs provide a more expressive model for processes evolving over discrete time points — a fully connected DBN with intra-time-slice arcs has more free parameters than the fully connected CTBN. (For example, if there are 2 binary variables, a fully-connected DBN has 12 free parameters and a fully-connected CTBN has only 8.) Thus, the DBN can represent certain transition models that do not arise from a purely Markovian continuous-time process. Overall, DBNs are a good choice for domains where the data is naturally time-sliced and where questions about events occuring between time points are not relevant. However, there are domains where the data has no natural timeslices (e.g., computer system monitoring or web/database transactions). Such domains are more naturally modelled as CTBNs than DBNs and the estimation of fewer parameters make CTBNs simpler to learn.

However, accurate modelling of most datasets will require an extension of the work presented here. The Markovian assumption restricts the expressive power of CTBNs to modeling exponential distributions over time. To allow more expressive distributions, hidden state must be introduced, either explicitly (through hidden variables) or implicitly (by using delayed exponentials or mixtures of exponentials). These extensions are not straightforward. Whereas in traditional Bayesian networks, a hidden variable takes any of a discrete number of possible values, in CTBNs a hidden variable takes any trajectory as a value. The space of trajectories is infinite both in the number of transitions and the times at which the transitions occur. Our current research is focused on adding hidden state which would make CTBN learning applicable to a wide range of practical applications where full observability is typically an unrealistic assumption.

**Acknowledgments.** This work was funded by ONR contract N00014-00-1-0637 under the MURI program "Decision Making under Uncertainty."

## References

Chickering, D. M., Geiger, D., & Heckerman, D. (1994). *Learning Bayesian Networks is NP-Hard* (Technical Report MSR-TR-94-17). Microsoft Research.

Dean, T., & Kanazawa, K. (1989). A model for reasoning about persistence and causation. *Computational Intelligence, 5*, 142–150.

Geiger, D., & Heckerman, D. (1995). *A characterization of the dirichlet distribution with application to learning Bayesian networks* (Technical Report MSR-TR-94-16). Microsoft Research.

Gihman, I. I., & Skorohod, A. V. (1973). *The theory of stochastic processes II*. New York: Springer-Verlag.

Heckerman, D., Geiger, D., & Chickering, D. M. (1995). Learning Bayesian networks: The combination of knowledge and statistical data. *Machine Learning, 20*, 197–243.

Lam, W., & Bacchus, F. (1994). Learning Bayesian belief networks: An approach based on the MDL principle. *Computational Intelligence, 10*, 269–293.

Nodelman, U., Shelton, C. R., & Koller, D. (2002). Continuous time Bayesian networks. *Proceedings of the Eighteenth Conference on Uncertainty in Artificial Intelligence* (pp. 378–387).